\documentclass[11pt,a4paper,logo]{googledeepmind}

\setleftlogo[120pt]{imgs/logo-removebg-preview-Photoroom.jpg} 
\setrightlogo[180pt]{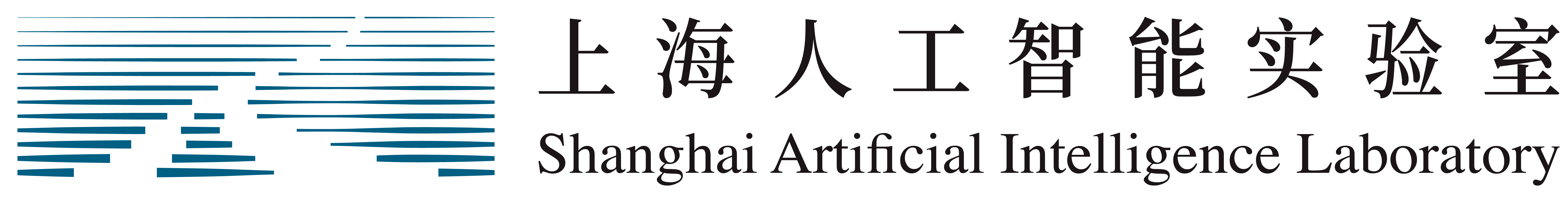}

\usepackage[T1]{fontenc}
\usepackage{pifont}
\usepackage[
    natbib=true,
    backend=biber,
    style=numeric, 
    sorting=none 
]{biblatex}
\AtEveryBibitem{\clearfield{month}}
\AtEveryBibitem{\clearfield{day}}
\usepackage{csquotes}

\newcommand{\sname}{AutoMLGen\xspace}

\title{\sname: Navigating Fine-Grained Optimization for Coding Agents}

\correspondingauthor{
$\spadesuit$: Co-first Author \quad
$\clubsuit$: Corresponding Author \\ 
Please send correspondence regarding this report to zhangbo@pjlab.org.cn
}

\author[1 2 $\spadesuit$]{Shangheng Du}
\author[1 $\spadesuit$]{Xiangchao Yan}
\author[1 $\spadesuit$]{Dengyang Jiang}
\author[1]{Jiakang Yuan}
\author[1]{Yusong Hu}
\author[1]{Xin Li}
\author[1 2]{Liang He}
\author[1 $\clubsuit$]{Bo Zhang}
\author[1 $\clubsuit$]{Lei Bai}

\affil[1]{Shanghai Artificial Intelligence Laboratory}
\affil[2]{East China Normal University}

\usepackage{pdflscape}

\usepackage{textcomp}
\usepackage{rotating}

\usepackage{setspace}
\usepackage{microtype} 

\usepackage{soul} 

\usepackage{graphicx}
\usepackage{subcaption}
\usepackage{caption}

\usepackage{booktabs}
\usepackage[table]{xcolor}
\sethlcolor{green!14}

\usepackage{array}
\usepackage{multirow}

\usepackage{amsmath}
\usepackage{siunitx}

\usepackage{enumitem}
\usepackage{float}
\usepackage{seqsplit}
\usepackage{framed}

\usepackage{tikz}
\usepackage{hyperref}
\usepackage{url}
\usepackage{listings}
\usepackage{xcolor}
\usepackage{soul}
\usepackage{colortbl}  
\usepackage{graphicx}
\usepackage{wrapfig}
\usepackage[most]{tcolorbox}
\usepackage{enumitem}
\usepackage{hyperref}

\tcbuselibrary{listingsutf8}
\definecolor{mycolor}{RGB}{50,80,150}

\usepackage{ragged2e}
\usepackage[most]{tcolorbox}

\usepackage{makecell}
\usepackage{adjustbox}

\usepackage[symbol]{footmisc}
\newcolumntype{Y}{>{\RaggedRight\arraybackslash}X}

\newcolumntype{C}{>{\centering\arraybackslash}X} 
\usepackage{colortbl}
\usepackage{xspace}
\usepackage{pifont}
\usepackage{bbding}
\usepackage{multirow}
\usepackage{booktabs,tabularx,xcolor}
\usepackage{enumitem}
\usepackage{ulem}
\usepackage{tabularx}
\usepackage{array}
\usepackage{wrapfig}
\usepackage{booktabs,tabularx,array}
\newcolumntype{L}{>{\raggedright\arraybackslash}X}
\newcolumntype{Y}{>{\centering\arraybackslash}X}  
\newcolumntype{Z}{>{\raggedleft\arraybackslash}X}  
\usepackage{textcomp}

\usepackage{hyperref}
\hypersetup{
    colorlinks=true,
    linkcolor=blue, 
    citecolor=blue,  
    filecolor=black,
    urlcolor=blue    
}

\usepackage{tocloft}

\usepackage{etoolbox}
\usepackage{arydshln}
\usepackage{ulem} 
\makeatletter
\patchcmd{\@tocline}
    {\hfil}
    {\leaders\hbox{\hfil}\hfil}
    {}{}
\makeatother

\begin{document}
\sloppy

\begin{abstract}

Large language models (LLMs) have shown impressive performance in general programming tasks. However, in Machine Learning Engineering (MLE) scenarios such as AutoML and Kaggle competitions, achieving high performance depends heavily on expert intervention and repeated adjustments rather than simply generating correct code. When applied directly to these tasks, LLMs often lack fine-grained domain priors, and existing MLE approaches that use linear or tree-structured searches limit knowledge transfer to adjacent hierarchical links. As a result, they cannot leverage past full trajectories or share information across branches, limiting self-evolving ability and search space diversity.
To address these limitations, we introduce \sname, an LLM-based coding agent that integrates a domain knowledge base for high-quality prior guidance and Monte Carlo Graph Search (MCGS) for efficient exploration. MCGS retains the tree-guided exploration of MCTS while embedding a graph structure into the expansion stage to enable dynamic path reorganization, historical trajectory reuse, and multi-solution fusion to support both self-evolution and collaborative learning. Combined with fine-grained operator sets, this design improves stability and accelerates convergence. 
Evaluation on the MLE-Bench shows that \sname achieves state-of-the-art performance in numerous dimensions, such as the average medal rate and the valid submission rate, under a 12-hour budget (half the standard runtime). The code is available at \url{https://github.com/Alpha-Innovator/InternAgent}.

\end{abstract}

\maketitle
\begin{figure}[thbp]  
\begin{center}
    \vspace{-0.5em}
    \includegraphics[width=0.94\linewidth]{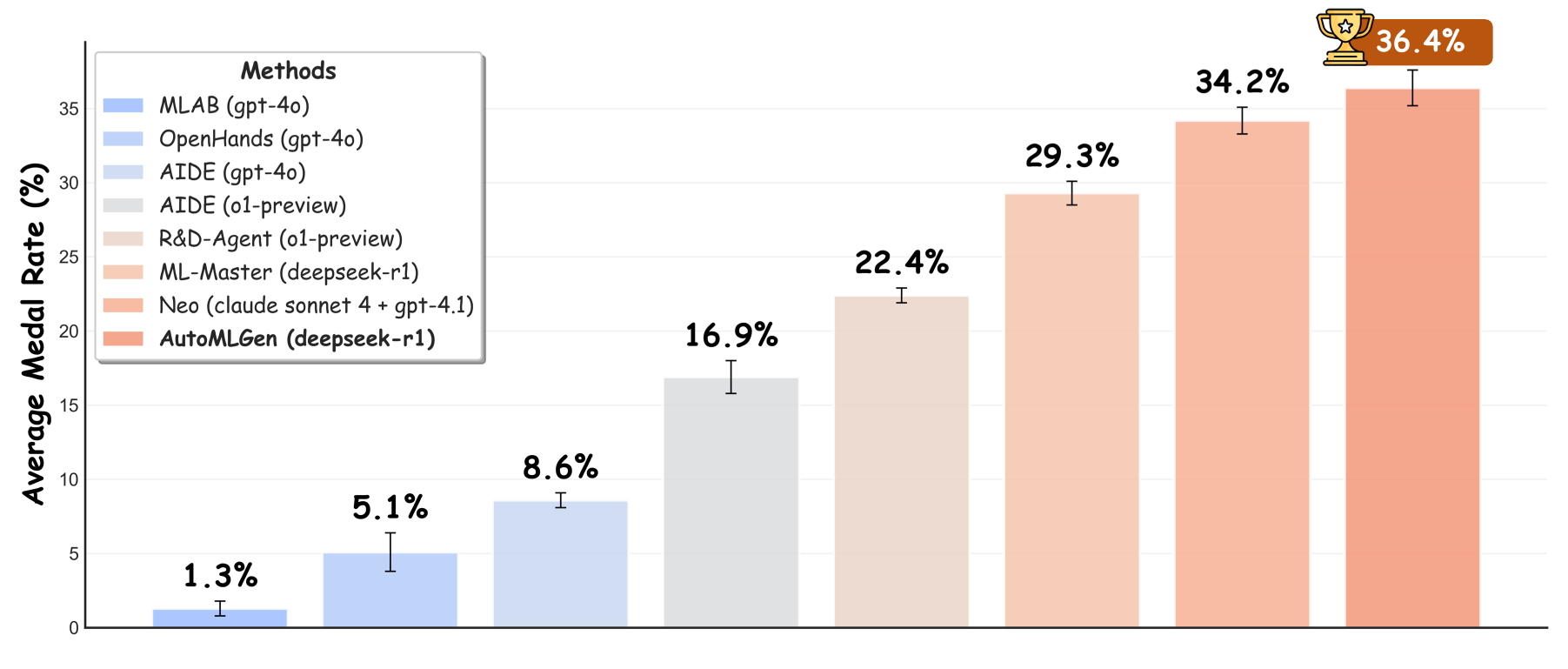}
    \vspace{-0.5em}
    \caption{\textbf{The comparison across various methods on MLE-Bench.} Our \sname wins the championship within a 12-hour budget.}
\end{center}
\end{figure}
\section{Introduction}

Benefiting from the increasing capability in coding and task planning, Large Language Models (LLMs)~\citep{gpt-4o,deepseek-r1} are shifting from simple code assistants~\citep{qian2024chatdev,hong2024metagpt} to autonomous agents of sophisticated Machine Learning Engineering (MLE)~\citep{amershi2019software-mle,mle-bench}. In the realm of MLE, LLM agents are required to enhance specific metrics for the given task by iteratively optimizing code, which requires a comprehensive consideration of various factors such as data, model architectures, and so on. While recent development of AutoML~\citep{he2021automl,feurer2022auto-sklearn} has brought about significant progress in optimizing discrete stages such as data processing, they often fall short of managing the entire end-to-end MLE workflow (\textit{i.e.}, from data preparation to model training and inference).

Recent advancements in AI agents, have spurred the creation of MLE agents~\citep{wang2024openhands,mlab,guo2024ds-agent}, which leverage the planning and execution capabilities of LLMs to optimize task performance across a broader search space. As a pioneer, AIDE~\citep{aide} reformulates the exploration process of optimizing codes as a tree search and achieves gold medals in some Kaggle competitions. R\&D-Agent~\citep{rdagent} iteratively refines codes through the cooperation of the researcher agent and the developer agent. ML-Master~\citep{ml-master} introduces a selectively
scoped memory mechanism and standard MCTS to integrate exploration and reasoning.

Despite the remarkable improvements on MLE tasks, existing MLE agents still suffer from the following issues. First, previous works exhibit an \textbf{\textit{over-reliance on the internal knowledge of LLMs}}. This dependence becomes a bottleneck when handling tasks in specialized domains where the internal knowledge of LLMs is often incomplete or absent. Consequently, the agent cannot integrate external domain expertise and optimize code effectively. Second, current MLE agents~\citep{aide,ml-master} mainly employ tree-structured search paradigms (\textit{e.g.}, MCTS), which may lead to \textbf{\textit{node isolation}}. This issue manifests in several ways: (1) Policy updates are driven primarily by feedback from immediate parent nodes, preventing the agent from abstracting the core reasons of success or failure across an entire trajectory. (2) Search proceeds in isolated branches, inhibiting the transfer and reuse of high-quality solutions discovered in one branch by others. (3) High-quality solutions are isolated in various branches, preventing their reorganization and integration into a better solution.

Motivated by this, we propose \textbf{\sname}, an LLM-based coding agent that integrates a curated ML knowledge base with \textbf{Monte Carlo Graph Search (MCGS)} algorithm for MLE tasks, automatically generating and refining ML pipelines through iterative exploration. Specifically, the knowledge base provides domain priors across model, data, and strategy dimensions, reducing cold start errors and supporting finer-grained improvements during search. To address the isolation and limited reuse in tree search, we introduce MCGS, a variant of MCTS that incorporates graph structure into the expansion stage, allowing trajectory recall, cross-branch reference, and multi-branch aggregation. In addition, a fine-grained operator set is designed to stabilize operations and improve executability.
Consequently, \sname achieves more stable and efficient exploration of end-to-end ML pipelines, leading to stronger solutions on challenging MLE tasks. Extensive experiments on MLE-Bench demonstrate its effectiveness, where \sname attains a 36.4\% average medal rate under a 12-hour budget, outperforming all existing baselines.

In conclusion, our key contribution can be summarized as follows:

\begin{itemize}[leftmargin=*]
    \item We propose \sname framework, the first graph-search-based end-to-end MLE task solver, which couples a curated domain knowledge base with MCGS to produce complete, high-quality ML pipelines by unifying general and specialized knowledge.
    
    \item We develop Monte Carlo Graph Search (MCGS), a variant of MCTS that introduces the compositional flexibility of graphs, thereby expanding search diversity and reusability. In addition, a set of fine-grained operators are designed to stabilize execution and enhance solution quality.
    
    \item Extensive experiments on MLE-Bench show that \sname achieves state-of-the-art performance under a 12-hour budget, including a 36.4\% average medal rate and 18.7\% gold medals, outperforming all existing baselines.

\end{itemize}
\section{Related Work}
\label{related_works}

\subsection{General-purpose Coding Frameworks}
Recent advances in Large Language Models have led to the development of powerful LLM-based agents~\citep{aider, wang2024openhands, wang2025repomaster} designed to tackle general software engineering tasks. Most early LLM-based agents were designed as general coding assistants, providing a flexible architecture without domain-specific tuning. For example, OpenHands~\citep{wang2024openhands} integrates LLM reasoning with tool use for complex software engineering tasks.  SWE-Agent~\citep{swe-agent} offers comprehensive command sets for navigating codebases and implementing solutions, achieving notable performance on software engineering benchmarks. Our work also aims to enhance the coding capabilities of LLM-based agents, but unlike these works, we focus on developing an advanced coding agent specially for ML task.


\subsection{Specialized Coding Agents for ML Engineering}
To address the unique challenges of machine learning engineering, a dedicated class of coding agents has been developed~\citep{aide,mle-star,mlzero,ml-master}, with many evaluated on comprehensive benchmarks like MLE-Bench~\citep{mle-bench}. These agents primarily frame the problem as a search for an optimal code-based solution. Early works like AIDE~\citep{aide} employ a greedy search strategy, which can be susceptible to local optima. To overcome this, subsequent frameworks have adopted more sophisticated exploration strategies. Multi-agent collaboration approaches like AutoKaggle~\citep{autokaggle} distribute tasks among specialized agents. Tree search has also emerged as a dominant paradigm. AutoMind~\citep{automind} introduces an agentic tree search grounded by an expert knowledge base, while R\&D-Agent~\citep{rdagent} manages parallel exploration traces. AI auto-research agents~\citep{team2025novelseek} systematically shows that high ML coding performance requires a careful co-design of both search policies and operators. However, these work often use isolated search paths and fail to facilitate the reuse of granular solutions. Our method resolves this inefficiency by fusing a knowledge base with MCGS to supplement task-specific knowledge and provide better recall, reference, as well as aggregation.



\begin{figure}[t]  
    \centering
    \vspace{-0.5em}
    \includegraphics[width=\linewidth]{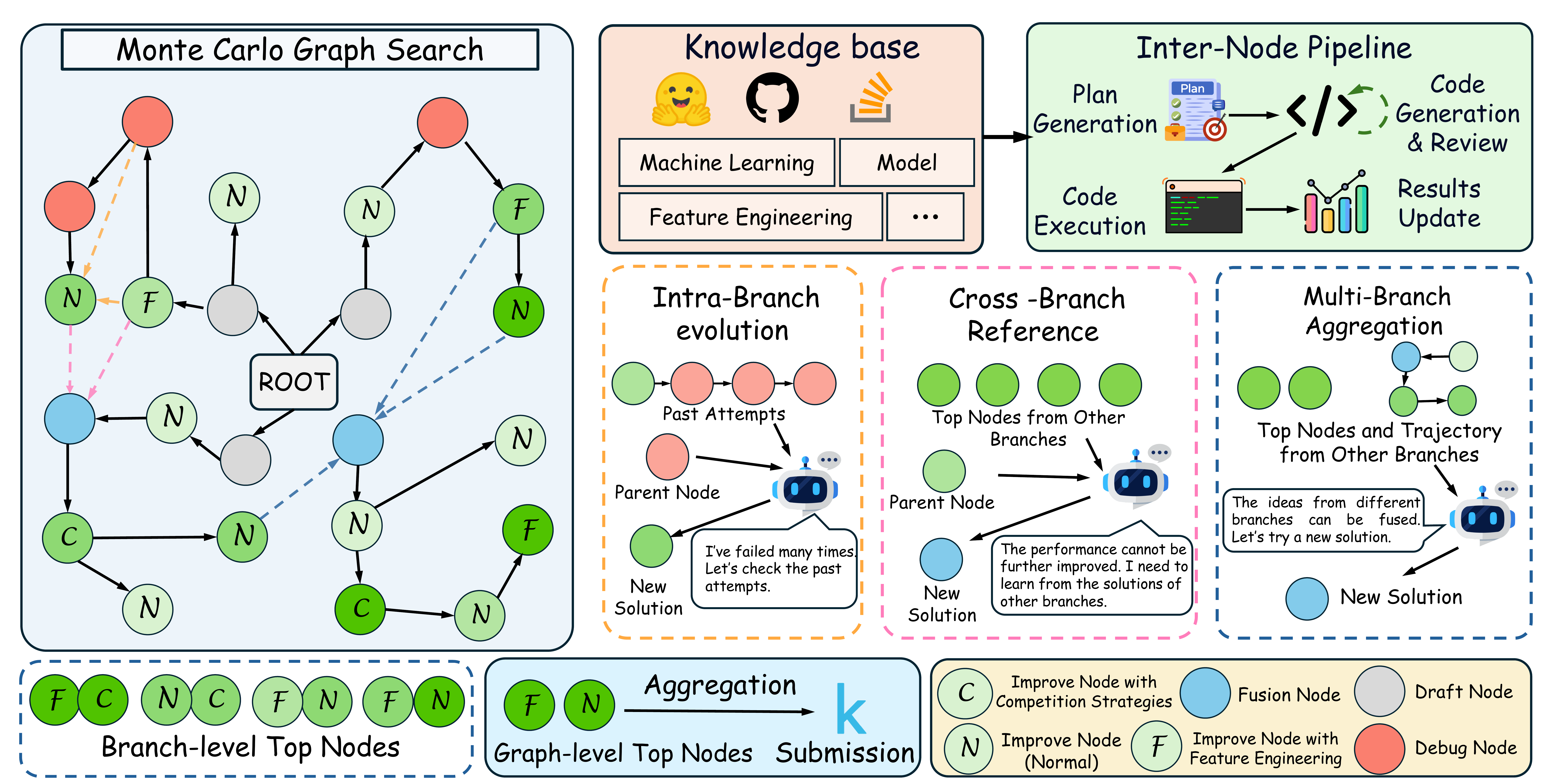}
    \caption{\textbf{The framework of \sname.} It consists of two main modules: (i) a curated ML domain knowledge base, and (ii) the MCGS module, which integrates graph-based exploration with a set of fine-grained operators. Detailed descriptions are provided in Section~\ref{sec:method}.}
    \vspace{-0.5em}
    \label{fig:framework}
\end{figure}

\section{\sname }
\label{sec:method}
In the MLE and automated algorithm design process, strong solutions often arise from careful design, reuse of past experience, and reference of multiple candidate pathways, rather than from a single linear refinement and iteration. Tree-based search methods \citep{ml-master,dojo}, such as MCTS, balance exploration and exploitation through branch-specific lineages, but this structure restricts knowledge flow and compositional reuse across branches and layers.


In this section, we introduce \textbf{\sname}, a framework for LLM-driven automatic ML pipeline generation, as shown in Figure~\ref{fig:framework}. The design combines three key components:
(1) a knowledge base that supplies ML domain priors and references for initialization and iterative search,
(2) MCGS, which extends MCTS-based pipeline with graph edges for trajectory reuse and cross-branch integration, and
(3) fine-grained operator sets that improve executability and stability.

\subsection{Problem Formulation}


Our objective is to automate the search, design, and optimization of end-to-end ML pipelines. We formalize the task as identifying the optimal solution within a search space~\citep{aide}, where each node represents a complete candidate pipeline spanning preprocessing, feature engineering, model training, and prediction. 
The goal is to select the best-performing solution for a given task:

\begin{equation}
\label{eq_1}
s^{*} = \arg\max_{s \in \mathcal{S}} h(T, s),
\end{equation}
where $h(T,s)$ denotes the evaluation of candidate solution $s$ on task $T$, which may vary by task (\textit{e.g.}, accuracy, AUC, or loss). The solution space $\mathcal{S}$, often organized as a tree or graph structure, contains all possible pipelines, and the search process aims to find the solution that optimizes the task metric.

\label{sec_know}
\subsection{ML Domain Knowledge Base}
Effective ML algorithm design typically relies on domain priors and hands-on experience. LLM knowledge alone is insufficient for complex tasks, leading to cold start and a high rate of early-stage errors. To address this, an ML domain knowledge base is curated and maintained, which improves the reliability of initial solutions, and provides ongoing reference during the search process.

\subsubsection{Knowledge Base Construction}

We design an ML domain knowledge base across three dimensions by synthesizing practices from open-source repositories and competition platforms such as Hugging Face, GitHub, followed by careful selection. \textbf{Model-level} knowledge categorizes models by application domain and provides concise descriptions with usage guidelines, enabling the agent to quickly select and operate suitable backbones across tasks. \textbf{Data-level} knowledge summarizes modality-specific constraints and preprocessing methods, highlights key feature-engineering principles. Finally, \textbf{strategy-level} knowledge focuses on practical tricks such as test-time augmentation (TTA) and ensembling methods, distilled from discussions of ML competitions.

\subsubsection{Knowledge Retrieval and Usage}

To preserve the agent’s autonomous exploration ability, model-level knowledge is injected only during initial solution generation. Given a task $T$, the system retrieves relevant elements $R_{KB}(T)$ by matching the task description with domain keywords such as audio, natural language processing, image classification). The retrieved knowledge includes concise model descriptions and usage guidelines, serving as lightweight priors to complement LLM reasoning. It is treated as an optional signal that the agent may adopt it, use it partially, or ignore it. Formally, the initial candidate is:
\begin{equation}
s_{init} = \mathrm{Init}(T, R_{KB}(T)),
\end{equation}

where $\mathrm{Init}$ denotes the initialization that the agent uses to generate plan and code.
During search, data- and strategy-level knowledge heuristically provides insight, enabling the agent to reason with more specific and advanced perspectives. 


\subsection{MCGS-guided Exploration in MLE}
In this section, we propose \textit{Monte Carlo Graph Search (MCGS)}, which extends MCTS by incorporating a graph structure into the expansion stage via branch–node dynamic fusion. MCGS explicitly introduces trajectory recall and branch-level experience aggregation, thereby enabling more flexible composition and improved knowledge sharing.

\subsubsection{Graph-based Search Space Formulation}
To realize the optimization objective in Equation~(\ref{eq_1}), we organize the search process over the solution space as a directed graph:
\begin{equation}
G = (V, E), \quad E = E_{T} \cup E_{\text{ref}},
\end{equation}
where the node set $V$ corresponds to candidate solutions, and each node $v \in V$ maps to a complete solution $s(v) \in \mathcal{S}$. Directed edges capture generative and reference relationships:

\begin{itemize}[leftmargin=*]
    \item \textbf{Primary edges} $E_{T}$: if $(u,v) \in E_{T}$, then node $v$ is obtained by applying an operator $o$ to node $u$ (i.e., $v = g_{o}(u)$). These edges preserve the parent–child generative order and are treated exactly as in classical MCTS statistics for selection and backpropagation.
    \item \textbf{Reference edges} $E_{\text{ref}}$: if $(r,v) \in E_{\text{ref}}$, then node $v$ obtains information from node $r$ as an extra reference beyond the parent link. Such edges connect nodes across branches or non-adjacent levels, enabling knowledge flow and compositional transfer, and they do not participate in backpropagation. When $E_{\text{ref}} = \varnothing$, the search reduces to standard tree-based MCTS.
\end{itemize}

\subsubsection{MCGS-based Exploration}

MCGS process follows the classical MCTS loop, retaining its strengths in selection and backpropagation, while extending the expansion phase with branch–node fusion in a dynamic graph. Through iterative exploration, the solution graph grows progressively to cover diverse candidate paths, and the best solution is returned at the stopping criterion.

\textbf{Selection.}
Although overall search space is formulated as a graph, the selection stage operates solely on the tree backbone formed by primary edges $E_T$. At the beginning of each iteration, the selection policy $\pi_{\text{sel}}$ traverses $E_T$ edges in a top-down manner to identify a node $v_t$ for expansion. For a given parent node $v$, the next child is chosen from its successors $\mathcal{C}(v)$ along $E_T$ using the UCT criterion:
\begin{equation}
\pi_{\text{sel}}(v) = \arg\max_{i \in \mathcal{C}(v)} \text{UCT}(i), 
\quad \text{where } \text{UCT}(i) = \frac{Q_i}{N_i + \varepsilon} + c \sqrt{\frac{\ln (N_v + 1)}{N_i + \varepsilon}},
\end{equation}

where $Q_i$ denotes the accumulated reward of child node $i$, $N_i$ is its visit count, $N_v$ is the visit count of the parent node $v$, and $c>0$ controls the strength of exploration, $\varepsilon > 0$ is a small smoothing constant to avoid division by zero. The selected node $v_t$ is then passed to expansion and evaluation.

\textbf{Expansion.}
To incorporate information flow and compositional reuse into the search process, we extend the original MCTS expansion with four types of operations:

\textbf{(1) Primary expansion.}
In this case, the new node is generated solely from its parent without referencing other nodes. 
Given the selected node $v_t$ and an operator $o \in \mathcal{O}$, expansion produces
\begin{equation}
v_{\text{new}} = g_{o}(v_{t}, \varnothing), 
\qquad (v_{t}, v_{\text{new}}) \in E_{T},
\end{equation}
where the reference set is empty ($R=\varnothing$), i.e., no cross-branch information is incorporated. This operation constitutes the baseline expansion, against which the graph-based variants extend. 
Typical operators in this form include, for example, \textit{Draft}, \textit{Improve}, and \textit{Debug}, as detailed in §\ref{sec_operators}.

\textbf{(2) Intra-branch evolution.}  
Inspired by human problem-solving strategies, this mode emphasizes reflecting on past attempts instead of blind trial and error. Practitioners review previous actions to see which changes improved outcomes or caused failures. Through self-reflection, the agent makes small adjustments, reinforcing effective patterns while avoiding repeated mistakes.  
Formally, given a node $v_{t}$, the agent takes the nearest $k$ nodes within the same branch to form a local trajectory, denoted as the intra-branch history reference set $\mathcal{R}_{\text{hist}}(v_{t}, k) \subseteq V$, and generates a new solution:

\begin{equation}
v_{\text{new}} = g_{o}(v_{t}, \mathcal{R}_{\text{hist}}(v_{t}, k)), 
\qquad (v_{t}, v_{\text{new}}) \in E_{T}, \;\;
\{(r, v_{\text{new}}) \mid r \in \mathcal{R}_{\text{hist}}(v_t, k)\} \subseteq E_{\text{ref}} .
\end{equation}

Here, $E_T$ preserves the parent–child relation, while $E_{\text{ref}}$ records the information flow from intra-branch history. The agent autonomously integrates both successful and failed experiences to form improved solutions, whereas selection and backpropagation are still conducted exclusively along $E_T$.

\textbf{(3) Cross-branch reference.} 
In ML competitions, contestants often draw inspiration from community-shared solutions when progress stalls. Similarly, MCGS selects a small set of high-quality nodes from other branches as references when the current branch stagnates.  
Formally, at a candidate node $v_{t}$, a reference set $\mathcal{R}_{\text{cross}}(N)$ is formed by taking the top-$N$ nodes across all evaluated branches, ranked by performance and stability. The new candidate is then generated as

\begin{equation}
v_{\text{new}} = g_{o}(v_{t}, \mathcal{R}_{\text{cross}}(N)), 
\qquad (v_{t}, v_{\text{new}}) \in E_{T}, \;\;
\{(r, v_{\text{new}}) \mid r \in \mathcal{R}_{\text{cross}}(N)\} \subseteq E_{\text{ref}} ,
\end{equation}

where $E_{\text{ref}}$ passes cross-branch knowledge, allowing agent to draw on strong solutions from other branches. Source selection and reuse are determined by the agent during candidate formation.

\textbf{(4) Multi-branch aggregation.}
For complex tasks, progress often requires synthesizing complementary insights from multiple strong solutions. This resembles a form of collective intelligence, where trajectories from different branches are merged and fragments of useful insights are combined to spark novel directions. When existing branches have accumulated sufficient trajectories, a new branch root is heuristically spawned beneath $v_{0}$, serving as a fresh starting point.
$\mathcal{R}_{\text{agg}} = \bigcup_{b \in \mathcal{B}} \mathcal{T}^{\text{top}}_{b}$ denote the reference set formed by aggregating top trajectories from multiple branches, where $\mathcal{T}^{\text{top}}_{b}$ represent the best-performing trajectories (or nodes) in branch $b$. A new candidate is generated as
\begin{equation}
v_{\text{new}} = g_{o}\!\big(v_{0}, \mathcal{R}_{\text{agg}}\big), 
\qquad (v_{0}, v_{\text{new}}) \in E_{T}, \;\;
\{(u, v_{\text{new}}) \mid u \in \mathcal{R}_{\text{agg}}\} \subseteq E_{\text{ref}} .
\end{equation}
Here, $E_{\text{ref}}$ records the knowledge sources being fused. Unlike incremental refinements along a single branch, this aggregation mechanism reorganizes thoughts from diverse origins into a wholly new branch, thereby opening an independent trajectory for exploration.

\textbf{Simulation.}
After generating a candidate $v_{\text{new}}$, its code is executed in an interpreter. The running outputs are parsed to extract the task-specific metric and the execution status and written back to the node.
Reward is computed relative to parent $v_{t}$, based on improvement: positive for higher scores, bonus for bug fixes, and penalties for failures or violations.

\textbf{Backpropagation.}
After simulation, reward and status are propagated to the root only along primary edges $E_{T}$, while reference edges $E_{\text{ref}}$ are excluded to keep credit assignment stable and interpretable. 
Each ancestor updates its visit count $N$ and value $Q$, guiding future UCT decisions. 
This shifts exploration toward promising trajectories, reducing dead ends and promoting stronger solutions.

\textbf{Memory Maintenance.} 
Throughout the search process, we maintain structured memory at three levels. 
At the node level, each node stores complete information, including its plan, code, metric, analysis, and state. 
At the branch level, we keep the top-$k$ nodes by metric, and at the graph level, the overall top-$k$ solutions are preserved until the end. 
This memory mechanism provides the basis for message passing across nodes and branches in our graph search space, while improving usability and interpretability during subsequent search and analysis.

\textbf{Parallelization.} 
Following R\&D-Agent, we extend MCGS with asynchronous branch-parallel exploration.
After expanding the root node $v_{0}$, multiple workers independently enter the selection stage and launch their own search traces, each proceeding with expansion and backpropagation in parallel.
Candidate code executions are also run in parallel threads, further improving resource utilization and accelerating discovery of diverse high-quality solutions.

\subsubsection{Finer-grained Operators}
\label{sec_operators}
Building on AIDE, a set of finer-grained operators are defined to support graph-based exploration.

\textbf{Draft.} This operator generates a solution from scratch, typically at initialization under the root or when new starting points are needed. Drafting may leverage the domain knowledge base (§\ref{sec_know}) for warm starts and reference existing memory to reduce duplication and enhance path diversity.  

\textbf{Debug.} This operator repairs faulty solutions when execution fails, guided by error traces (\textit{e.g.}, missing dependencies, tensor shape mismatches). It applies minimal modifications to restore executability.

\textbf{Improve.} This operator family refines executable solutions to achieve further performance gains while preserving executability. It comprises three variants: \textbf{Improve-Normal}, which applies small adjustments such as switching optimizers or hyperparameter changes; \textbf{Improve-FE} (Feature Enhancement), which emphasizes data augmentation and feature engineering (\textit{e.g.}, categorical encodings, feature aggregation); and \textbf{Improve-CS} (Competition Strategies), which introduces competition-style practices from the knowledge base (\textit{e.g.}, pseudo-labeling, ensembling).

\textbf{Fusion.} Triggered when a branch’s performance stalls or when the global structure stabilizes. This operator merges information from multiple candidate solutions by combining primary and reference edges, leveraging historical trajectory review and branch-level experience pooling to realize self-evolution and collective intelligence.

\textbf{Code Review.} After code generation, a reviewing operator checks for data leakage, naming or import errors, and metric–task mismatches. This helps maintain node quality and prevents overfitting.

\textbf{Ensemble.} During search, a global Top-$K$ set of candidate nodes is maintained. Near termination, the best solutions are heuristically combined to produce a more robust final solution.

\section{Experiments}

\subsection{Experiment Setup}

\textbf{Benchmark.} All experiments are tested on MLE-Bench~\citep{mle-bench}, a comprehensive benchmark introduced by OpenAI for evaluating how well AI agents perform at machine learning engineering. The full set of the MLE-Bench comprises 75 Kaggle tasks, categorized by complexity into low, medium, and high, while MLE-Bench Lite consists of a subset of 22 low-complexity tasks for teams with limited computational resources. More details are provided in Appendix~\ref{appen:benchmark}.

\textbf{Implementation details.} We adopt DeepSeek-R1-0528~\citep{deepseek-r1} to generate plans and Python code with temperature set to 0.5. For MCGS, the simulation budget is fixed at 500 steps and the UCT exploration constant is 1.414. 
For the single-task test environment, we use 32 Intel(R) Xeon(R) vCPUs, 230GB of RAM, and 1 NVIDIA A800 GPU with a 12-hour time budget and averaged results over 3 random seeds. More implementation details are introduced in Appendix~\ref{appeddix:hyper}
\begin{table}[!t]
\centering
\caption{Percentage of achieving any medals across different ML task complexity levels (left) and other evaluation dimensions (right) on MLE-Bench. Reporting results are mean ± SEM over 3 seeds; * denotes single run. Valid, Median+, and Gold indicate the percentage of submissions with valid, above median score, and gold medal; Best performances are marked in \textbf{bold}.}
\label{tab:mian_res}

\setlength{\tabcolsep}{4pt}        
\setlength{\aboverulesep}{1.2ex} 
\setlength{\belowrulesep}{0.9ex} 

\renewcommand{\arraystretch}{1.1} 

\small
\begin{tabularx}{1\linewidth}{@{}lCCCCC|CCC@{}} 
\hline
& & \multicolumn{4}{c|}{\textbf{Medal rate in different complexity}} &
\multicolumn{3}{c}{\textbf{Other evaluation dimensions}}\\
\cline{3-6}
\cline{7-9}
\textbf{Agent} & \textbf{Time} &\textbf{Low} & \textbf{Medium} & \textbf{High} & \textbf{Avg} & \textbf{Valid} & \textbf{Median+} & \textbf{Gold} \\
& (h) & (\%) & (\%) & (\%) & (\%) & (\%) & (\%) & (\%)\\
\hline

\multicolumn{9}{@{}l}{\textbf{MLAB}} \\
gpt-4o-24-08 & {24}& 4.2±1.5 & 0.0±0.0 & 0.0±0.0 & 1.3±0.5 & 44.3±2.6 & 1.9±0.7 & 0.8±0.5 \\
\arrayrulecolor{black!40}\hline

\multicolumn{9}{@{}l}{\textbf{OpenHands}} \\
gpt-4o-24-08 & {24}& 11.5±3.4 & 2.2±1.3 & 1.9±1.9 & 5.1±1.3 & 52.0±3.3 & 7.1±1.7 & 2.7±1.1 \\
\arrayrulecolor{black!40}\hline

\multicolumn{9}{@{}l}{\textbf{AIDE}} \\
gpt-4o-24-08 & {24}& 19.0±1.3 & 3.2±0.5 & 5.6±1.0 & 8.6±0.5 & 54.9±1.0 & 14.4±0.7 & 5.0±0.4 \\
o1-preview & {24}& 34.3±2.4 & 8.8±1.1 & 10.0±1.9 & 16.9±1.1 & 82.8±1.1 & 29.4±1.3 & 9.4±0.8 \\
Deepseek-R1* & {24}& 27.3±0.0 & 7.9±0.0 & 13.3±0.0 & 14.7±0.0 & 78.6±0.0 & 34.6±0.0 & 8.0±0.0 \\
\arrayrulecolor{black!40}\hline

\multicolumn{9}{@{}l}{\textbf{R\&D-Agent}} \\
o1-preview & {24}& 48.2±1.1 & 8.9±1.0 & 18.7±1.3 & 22.4±0.5 & 86.1±1.1 & 32.8±1.2 & 14.4±0.5 \\
\arrayrulecolor{black!40}\hline

\multicolumn{9}{@{}l}{\textbf{ML-Master}} \\
Deepseek-R1 & {12}& 48.5±1.5 & 20.2±2.3 & 24.4±2.2 & 29.3±0.8 & 93.3±1.3 & 44.9±1.2 & 17.3±0.8 \\
\arrayrulecolor{black!40}\hline

\multicolumn{9}{@{}l}{\textbf{Neo}} \\
\multicolumn{9}{@{}l}{Claude-Sonnet 4} \\
+ GPT-4.1 & {36}& 48.5±1.5 & \textbf{29.8±2.3} & 24.4±2.2 & 34.2±0.9 & 85.8±2.2 & 40.0±0.8 & 13.8±1.8 \\
\arrayrulecolor{black!40}\hline

\multicolumn{9}{@{}l}{\textbf{\sname\ (ours)}} \\
\rowcolor[RGB]{222,236,215}
Deepseek-R1 & \textbf{12} & \textbf{62.1±3.0} & 26.3±2.6 & \textbf{24.4±2.2} & \textbf{36.4±1.2} & \textbf{96.4±0.4} & \textbf{48.4±1.2} & \textbf{18.7±0.8} \\
\arrayrulecolor{black}\hline
\end{tabularx}

\end{table}


\noindent\textbf{Methods for comparison.} To provide a comprehensive comparison, we evaluate \sname alongside both methods tested on the full set of MLE-Bench and those only tested on MLE-Bench-Lite.
These include MLAB~\citep{mlab}, OpenHands~\citep{wang2024openhands}, AIDE~\citep{aide}, R\&D-Agent~\citep{rdagent}, ML-Master~\citep{ml-master}, Neo~\citep{neo}, MLE-Star~\citep{mle-star}, MLZero~\citep{mlzero}, KompeteAI~\citep{KompeteAI}, and AIRA-dojo~\citep{dojo}. We use results reported in MLE-Bench leaderboard or their paper.

\subsection{Main Results}

\begin{figure}[t]
  \centering
  \vspace{-1em}
  \begin{minipage}[b]{0.48\linewidth}
    \centering
    \includegraphics[width=\linewidth]{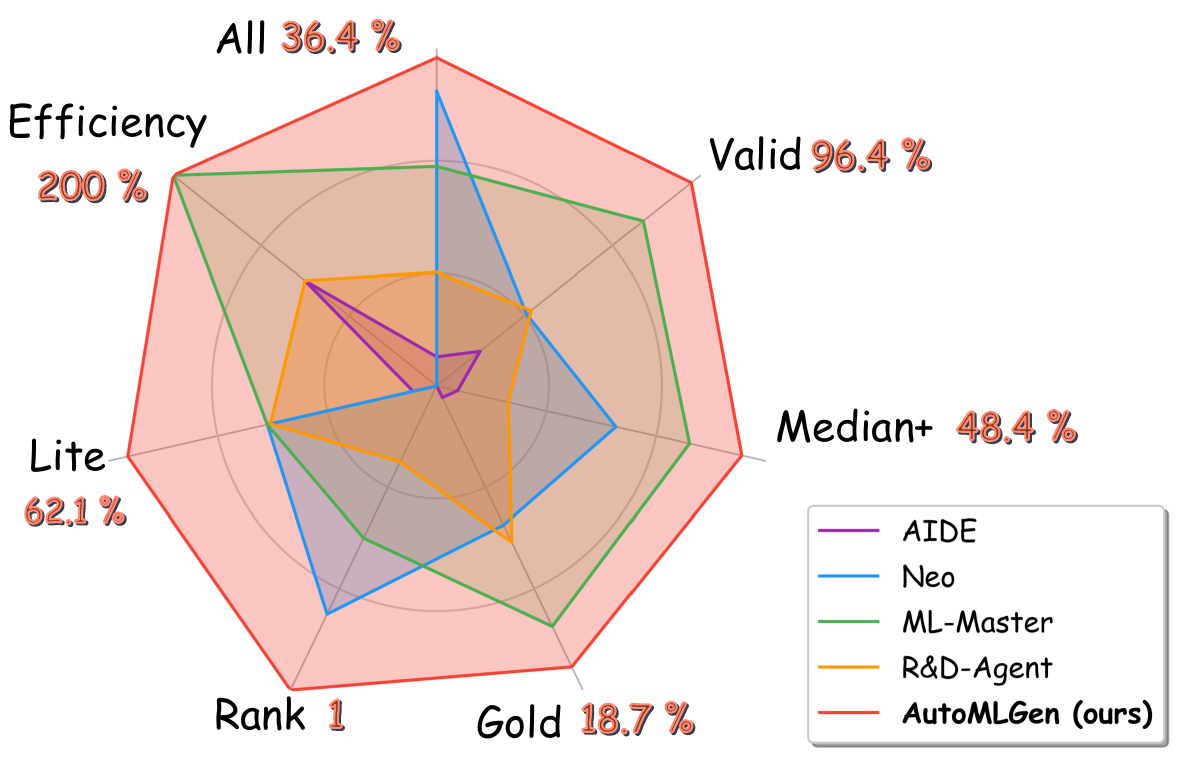}
    \caption{\textbf{MLE-Bench results of \sname and other methods.} It is noticeable that \sname performs better at all these dimensions with the shortest run time.}%
    \label{fig:radar}
  \end{minipage}%
  \hfill
  \begin{minipage}[b]{0.5\linewidth}
    \centering
    \includegraphics[width=\linewidth]{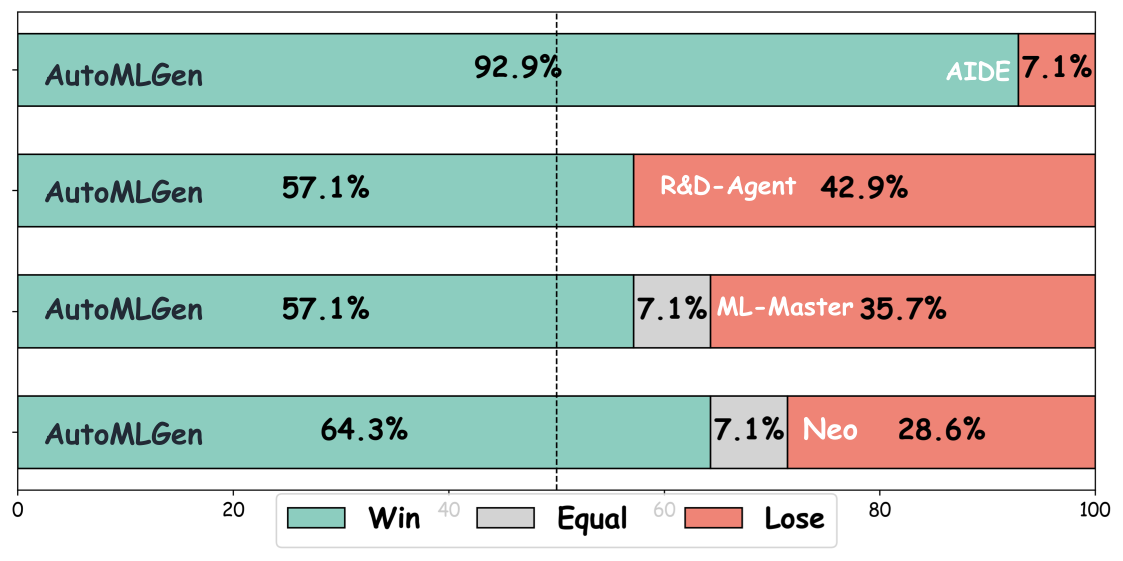}
    \caption{\textbf{Performance comparison on all high-level tasks of MLE-Bench.} Win means the average (3 seeds) test score of the task is better, so on for equal and lose. It can be seen that \sname achieves a better win rate against different baselines.}%
    \label{fig:high}
  \end{minipage}
\end{figure}

\noindent\textbf{\sname achieves state-of-the-art performance across MLE-Bench.} As demonstrated in Table~\ref{tab:mian_res} and Figure~\ref{fig:radar}, our proposed method, \sname, achieves superior performance compared to all baseline methods. Notably, \sname achieves an average medal rate of 36.4\%  and an impressive gold medal rate of 18.7\%, which are the highest among all evaluated approaches. These results highlight the robustness of \sname across varying levels of task complexity. Specifically, \sname outperforms the second-best method by a significant margin in the low-complexity category (62.1\% vs. 48.5\%) and the score improvement in complex categories (analyzing in detail later), demonstrating its adaptability to diverse ML challenges.
In addition to other evaluation dimensions, \sname achieves the highest valid submission rate of 96.4\%, indicating its reliability in producing consistently valid results. Furthermore, \sname surpasses human-level performance in 48.4\% of tasks, further demonstrating its ability to generalize effectively across diverse scenarios. Compared to Neo~\citep{neo}, the second-best approach, \sname not only demonstrates higher medal rates but also achieves these results with reduced time consumption and computational cost. For instance, while Neo requires 36 hours to achieve its performance, \sname achieves superior results with only 12 hours of computation time, emphasizing its efficiency and scalability.
When compared with the methods tested only on MLE-Bench-Lite, \sname similarly achieves state-of-the-art performance (Table~\ref{tab:lite}), further solidifying its position as a leading method. These results collectively highlight \sname's exceptional performance, efficiency, and robustness across diverse ML tasks, setting a new standard for future benchmarks and evaluations.

\begin{wraptable}{l}{0.6\textwidth}  
  \centering
  \caption{Performance comparison on MLE-Bench-Lite.\\ * means single run. Best performances are marked in \textbf{bold}.}
  \label{tab:lite}
  \footnotesize      
  \begin{tabular}{lc}
    \toprule
    \textbf{Agent} & Medal Rate (\%) \\
    \midrule
    MLZero* (Claude-Sonnet 3.7) & 36.4 \\
    MLE-Star (Gemini-2.0-flash)  & 43.9$\pm$6.2 \\
    AIRA-dojo* (o3)              & 47.7 \\
    KompeteAI (gemini-2.5-flash) & 51.5$\pm$1.5 \\
    \rowcolor[RGB]{222,236,215}
    \textbf{\sname} (Deepseek-R1) & \textbf{62.1$\pm$3.0} \\
    \bottomrule
  \end{tabular}
\end{wraptable}

\noindent\textbf{\sname demonstrates a stronger ability to handle more complex problems.} In the high-level tasks of MLE-Bench, although \sname achieves an equivalent medal rate to the other two top-performing candidates~\cite{ml-master,neo}, as shown in Table~\ref{tab:mian_res}, a deeper analysis of the average task scores, illustrated in Figure~\ref{fig:high}, reveals that our method consistently outperforms the baselines across a larger number of tasks. This highlights the robustness and versatility of \sname when addressing the most challenging ML tasks. The higher overall scores, despite similar medal rates, indicate that the finer-grained optimization of our framework yields more stable and consistent improvements even in difficult scenarios.

\subsection{Ablation Study and Analysis}
\textbf{Ablations on proposed components.} We conduct ablation experiments on MLE-Bench-Lite with a single seed run to evaluate the effectiveness of the proposed modules (Table~\ref{tab:ablation_seed3}). The \textit{baseline} is a standard MCTS-based agent without external knowledge or graph extensions. We first add the ML domain knowledge base improves the medal rate from 40.91\% to 50.00\%, indicating that domain priors reduce cold-start errors and guide finer refinements. Building on this, applying intra-branch evolution of MCGS as reference edges leverages historical trajectories within the same branch, further boosting the medal rate to 59.09\%. Finally, the complete framework is realized by merging cross-branch references and multi-branch aggregation, achieving a 68.12\% medal rate, which demonstrates the value of reusing and reorganizing high-quality components across branches to promote both diversity and stability. Overall, the ablation results highlight that each component contributes to the \sname framework’s ability to handle ML tasks.

\begin{figure}[t]  
    \centering
    \includegraphics[width=\linewidth]{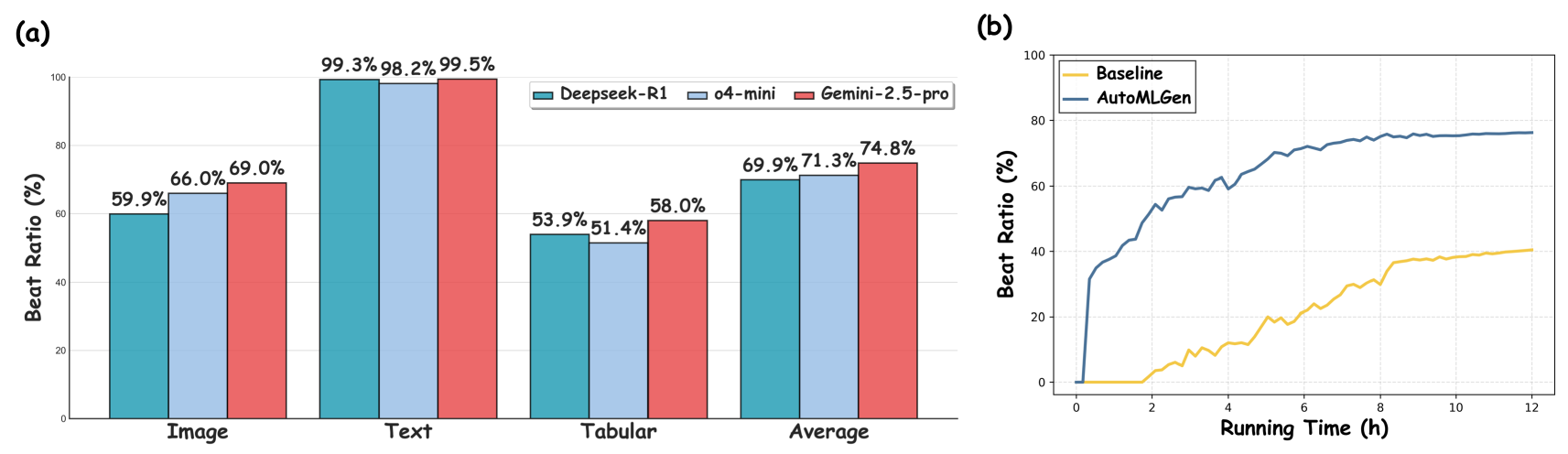}
    \caption{\textbf{(a) The comparison of different models by task type}. \sname is tested with different LLMs (DeepSeek-R1, o4-mini, and Gemini-2.5-pro) across image, text, and tabular tasks. 
\textbf{(b) The evolution of beat ratio over time}. This figure shows how \sname compares with the baseline under a 12-hour budget, where \sname consistently achieves higher leaderboard standings as search progresses.}


\label{fig:analysis}
\end{figure}

\textbf{Performance with different LLMs.} We also evaluate \sname across three state-of-the-art LLMs on a subset of MLE-Bench tasks: DeepSeek-R1~\citep{deepseek-r1}, o4-mini~\citep{o4-mini}, and Gemini-2.5-pro~\citep{gemini2.5}. As shown in Figure~\ref{fig:analysis} (a), all models achieve comparable performance in text processing tasks, while showing greater variation in image and tabular domains. DeepSeek-R1 and o4-mini demonstrate similar overall performance, with Gemini-2.5-pro achieving the highest average performance. These results indicate that \sname scales with underlying model capacity and remains adaptable across distinct foundation models. More detailed results can be found in Appendix~\ref{appendix:diff_res}.

\textbf{Performance over time.} To analyze the trend in the performance of \sname over time, we conducted an evaluation  of Beat ratio vs. runtime which is presented in Figure~\ref{fig:analysis} (b). As illustrated in the figure, the performance of our method improves progressively with increasing running time, which can be attributed to the proposed MCGS module's ability to interact with same/cross branch and effectively aggregate those. Furthermore, at each time step, our method consistently outperforms the baseline, demonstrating the effectiveness of the proposed components.

\begin{table}[t]
\centering
\caption{Ablation study on knowledge base and MCGS on MLE-Bench-Lite.}
\label{tab:ablation_seed3}

\setlength{\tabcolsep}{5pt}
\renewcommand{\arraystretch}{1.12}
\footnotesize
\resizebox{0.9\linewidth}{!}{
\begin{tabular}{lccc} 
\toprule
\textbf{Methods} & \textbf{Medal (\%)} & \textbf{ Median+ (\%)} & \textbf{Beat (\%)} \\
\midrule
baseline          & 40.91\% & 68.18\% & 65.33\% \\
+ knowledge base       & 50.00\% & 77.27\% & 68.59\% \\
+ knowledge base + MCGS (only Intra-branch)    & 59.09\% & 81.82\% & 73.20\% \\
\midrule
\rowcolor[RGB]{222,236,215}
\textbf{\sname(+ knowledge base + MCGS )}              & \textbf{68.12\%} & \textbf{86.36\%} & \textbf{78.33\%} \\
\bottomrule
\end{tabular}
}
\end{table}

\section{Conclusion and Discussion}
In this paper, we present \sname, an LLM-based agent that combines a curated ML knowledge base with Monte Carlo Graph Search (MCGS) to address key limitations of current MLE approaches. 
The knowledge base provides domain priors across model, data, and strategy dimensions, improving cold-start performance and guiding finer-grained refinements. 
MCGS transforms the tree-structured search space into a graph, introducing trajectory recall and branch-level aggregation to support self-evolving and collective intelligence. 
Together with a set of specialized operators, these components enable more stable, efficient, and diverse exploration of end-to-end ML pipelines. 
Evaluation on MLE-Bench shows that \sname achieves 36.4\% average medal rate under only a 12-hour budget, outperforming all existing baselines. Additional experiments further confirm the effectiveness of MCGS and the curated knowledge base across diverse tasks. 
In the future, we will extend \sname to broader benchmarks beyond MLE-Bench and incorporate multi-step, decomposed code generation to handle more complex AI tasks.

\begingroup
\sloppy
\normalem
\printbibliography[heading=bibintoc]
\endgroup

\clearpage
\appendix
\section{Appendix}
\subsection*{Use of LLMs}
We use large language models (LLMs) only to assist in drafting and refining our manuscripts, helping improve clarity and coherence. 

\subsection{MLE-Bench Benchmark}
\label{appen:benchmark}
Machine Learning Engineering (MLE) represents a critical frontier in AI development, requiring sophisticated integration of coding, experimentation, and problem-solving skills. Researchers usually evaluate such capacity of an LLM agent on MLE-bench proposed by OpenAI. 

Our work is also carried out on this benchmark. We now introduce MLE-Bench in detail:

MLE-bench is a comprehensive benchmark designed to assess autonomous ML engineering performance through real-world competitions.
It comprises 75 carefully curated Kaggle competitions spanning diverse domains, including natural language processing, computer vision, signal processing, and tabular data analysis. These competitions are selected from 586 candidates through rigorous manual screening by ML engineers, ensuring each task represents authentic, challenging ML engineering work relevant to contemporary practice. The dataset includes competitions of varying complexity: 22 low-complexity tasks (solvable by experienced engineers in under 2 hours), 38 medium-complexity tasks (2-10 hours), and 15 high-complexity tasks (over 10 hours), covering 15 distinct problem categories.
Each competition in MLE-bench includes the original problem description, datasets with reconstructed train-test splits, local grading code, and human baseline performance from Kaggle leaderboards. This setup enables direct comparison between AI agents and human competitors while maintaining evaluation integrity. The benchmark employs medal achievement rates as the primary metric, where agents must reach bronze, silver, or gold medal thresholds based on their performance relative to human participants.
The benchmark evaluates end-to-end ML engineering capabilities, including data preprocessing, model architecture design, hyperparameter tuning, training optimization, and debugging. Agents must work autonomously within time constraints (24-hour time limit) to produce valid submission files. This comprehensive evaluation framework reveals both the promise and limitations of current AI systems in performing complex ML engineering tasks, providing crucial insights for the development of more capable autonomous ML systems.

\subsection{Metric for Evaluation}
\label{appen:eval}
In this section, we introduce the key metrics used to assess the performance of our agent. These metrics are similar to those used by humans in Kaggle competitions. Each metric we used in the main paper is summarized below:
 
\begin{itemize}[leftmargin=0.3in]
    \item \textbf{Average Medal Rate (Avg, in $\%$):} represents the average number of task submissions that can win the medal, including silver, bronze, and gold. The threshold for the score that can earn a medal is officially provided by Kaggle and MLE-Bench.
     \item \textbf{Valid Submission Rate (Valid, in $\%$):} represents validity rate of the submitted results. The submission format and other validity checks are officially provided by Kaggle and MLE-Bench.
     \item \textbf{Above Median Rate (Median+, in $\%$):} represents the average number of task submissions that can beat half of the human competitors. The threshold for the score that can beat half of the human competitors is officially provided by Kaggle and MLE-Bench.
     \item \textbf{Gold Medal Rate (Gold, in $\%$):} represents the average number of task submissions that can win the gold medal. The threshold for the score that can earn  the gold medal is officially provided by Kaggle and MLE-Bench.
     \item \textbf{Agent  Runtime (Time in $\%$ or Efficiency):} represents the work time for agents to produce submission files. Less running time means higher efficiency. 
     \item \textbf{Above Beat Ratio (Average Beat, in $\%$):}  represents the average percentage of human competitors whose performance is surpassed by the task submission results.
. The top percentage of each score for the contestants (i.e., the beat ratio) is officially provided by Kaggle and MLE-Bench.
\end{itemize}

\subsection{Hyperparameters}
We provide the default hyperparameter configuration used in our MCGS framework (Table~\ref{tab:mcgs_hparams}). These hyperparameters are used throughout all experiments unless otherwise specified, and can be tuned to adapt the algorithm to different domains or computational budgets.
\label{appeddix:hyper}
\begin{table}[t]
\centering
\caption{MCGS Hyperparameter Configuration.}
\label{tab:mcgs_hparams}

\setlength{\tabcolsep}{4pt}
\renewcommand{\arraystretch}{1.12}
\footnotesize

\begin{tabularx}{0.9\linewidth}{@{}l L r@{}}
\toprule
\textbf{Hyperparameter} & \textbf{Description} & \textbf{Default} \\
\midrule

\multicolumn{3}{@{}l}{\textit{General Search}}\\[-2pt]
\arrayrulecolor{black!30}\cmidrule(lr){1-2}\arrayrulecolor{black}
max\_steps            & Max search steps   & 500 \\
exploration\_constant & UCT exploration constant \(C\)             & 1.414 \\
temperature           & LLM decoding temperature                   & 0.5 \\
max\_parallel\_workers& Max parallel workers                   & 3 \\
max\_draft\_num       & Max Draft attempts from root                & 7 \\
max\_debug\_num       & Max Debug attempts                          & 20 \\

\midrule
\multicolumn{3}{@{}l}{\textit{Memory}}\\[-2pt]
\arrayrulecolor{black!30}\cmidrule(lr){1-2}\arrayrulecolor{black}
branch\_top\_k        & Top-$k$ candidates kept per branch          & 5 \\
global\_top\_k        & Top-$k$ solutions kept globally             & 10 \\

\midrule
\multicolumn{3}{@{}l}{\textit{Reference / Fusion}}\\[-2pt]
\arrayrulecolor{black!30}\cmidrule(lr){1-2}\arrayrulecolor{black}
max\_history\_num                & Max historical trajectories used in intra-branch  & 7 \\
max\_ref\_num       & Max reference solutions used in cross-branch  & 7 \\
max\_agg\_num       & Max aggregation trajectories used in multi-branch  & 7 \\
ensemble\_num           & Final ensemble size                                & 6 \\

\midrule
\multicolumn{3}{@{}l}{\textit{Knowledge base}}\\[-2pt]
\arrayrulecolor{black!30}\cmidrule(lr){1-2}\arrayrulecolor{black}
kb\_init\_ref\_prob   & Heuristic probability of KB reference at initialization & 0.8 \\

\bottomrule
\end{tabularx}
\end{table}

\subsection{More detailed results of different LLMs}
\label{appendix:diff_res}
\begin{table*}[t]
  \centering
  \caption{Score comparison on 10 MLE-Bench tasks.  
           Best result for each task is highlighted in \textbf{bold}.}
  \label{tab:selected_10_tasks_score}
  \resizebox{1.0\linewidth}{!}{
  \setlength{\tabcolsep}{6pt}
  \begin{tabular}{llccc}
    \toprule
    \textbf{Task} & \textbf{Metric} & \textbf{DeepSeek-R1} & \textbf{o4-mini} & \textbf{Gemini-2.5-pro} \\
    \cmidrule(lr){3-5}
    \multicolumn{5}{l}{\textit{Image Tasks}}\\
    \midrule
    dog-breed-identification                 & Logloss $\downarrow$  & \textbf{0.3003} & 0.3941 & 0.3418 \\
    ranzcr-clip-catheter-line-classification & AUC $\uparrow$        & 0.9162 & 0.9040 & \textbf{0.9403} \\
    histopathologic-cancer-detection         & AUC $\uparrow$        & \textbf{0.9981} & 0.9940 & 0.9980 \\
    denoising-dirty-documents                & RMSE $\downarrow$     & 0.0418 & 0.0181 & \textbf{0.0165} \\
    \midrule
    \multicolumn{5}{l}{\textit{Text Tasks}}\\
    \midrule
    jigsaw-toxic-comment-classification      & AUC $\uparrow$        & 0.9873 & 0.9869 & \textbf{0.9879} \\
    spooky-author-identification             & Logloss $\downarrow$  & 0.2163 & 0.2534 & \textbf{0.2113} \\
    detecting-insults-in-social-commentary   & Accuracy $\uparrow$   & 0.9391 & 0.9388 & \textbf{0.9470} \\
    \midrule
    \multicolumn{5}{l}{\textit{Tabular Tasks}}\\
    \midrule
    new-york-city-taxi-fare-prediction       & RMSE $\downarrow$     & 5.7589 & 6.2157 & \textbf{4.6956} \\
    nomad2018-predict-transparent-conductors & RMSLE $\downarrow$    & \textbf{0.0585} & 0.0591 & 0.0593 \\
    tabular-playground-series-may-2022       & Accuracy $\uparrow$   & \textbf{0.9796} & 0.9690 & 0.9793 \\
    \bottomrule
    
  \end{tabular}
  }
\end{table*}

To provide a comprehensive evaluation of \sname's adaptability across different foundation models, we conducted experiments using three state-of-the-art LLMs: DeepSeek-R1~\citep{deepseek-r1}, o4-mini~\citep{o4-mini}, and Gemini-2.5-pro~\citep{gemini2.5}. We selected a representative subset of 10 tasks from MLE-Bench, covering three distinct domains: image processing, text analysis, and tabular data tasks. Table~\ref{tab:selected_10_tasks_score} presents the detailed performance comparison across all three models on these selected tasks. The results reveal several interesting patterns:

\textbf{Image Tasks:} The performance varies significantly across models in image-related tasks. DeepSeek-R1 achieves the best performance on dog breed identification (Logloss: 0.3003) and histopathologic cancer detection (AUC: 0.9981), while Gemini-2.5-pro excels in catheter line classification (AUC: 0.9403) and document denoising (RMSE: 0.0165). This variation suggests that different LLMs may generate distinct approaches or architectures for computer vision problems, leading to varying effectiveness in the resulting ML solutions.

\textbf{Text Tasks:} All three models demonstrate remarkably consistent performance in text processing tasks, with minimal differences in scores. Gemini-2.5-pro slightly outperforms others across all text tasks, achieving the best results in toxic comment classification (AUC: 0.9879), author identification (Logloss: 0.2113), and insult detection (Accuracy: 0.9470). The small performance gaps indicate that all three LLMs possess strong capabilities in generating effective NLP solutions, likely due to their inherent understanding of text processing methodologies.

\textbf{Tabular Tasks:} Similar to image tasks, tabular data processing shows notable performance variations. Gemini-2.5-pro demonstrates superior performance in taxi fare prediction (RMSE: 4.6956), while DeepSeek-R1 achieves the best results in material property prediction (RMSLE: 0.0585) and the playground series classification task (Accuracy: 0.9796). These differences may reflect varying approaches to feature engineering, model selection, or hyperparameter optimization generated by different LLMs.

These results confirm that \sname is successfully adapting to different LLMs as backends for generation. The consistent performance across text tasks and the model-specific advantages in the image and tabular domains demonstrate that different LLMs bring their unique problem solving approaches to the generation of automated machine learning solutions, while \sname effectively harnesses these diverse capabilities.

\subsection{Case Study}
\begin{figure}[thbp]  
\begin{center}
    \vspace{-0.5em}
    \includegraphics[width=1.0\linewidth]{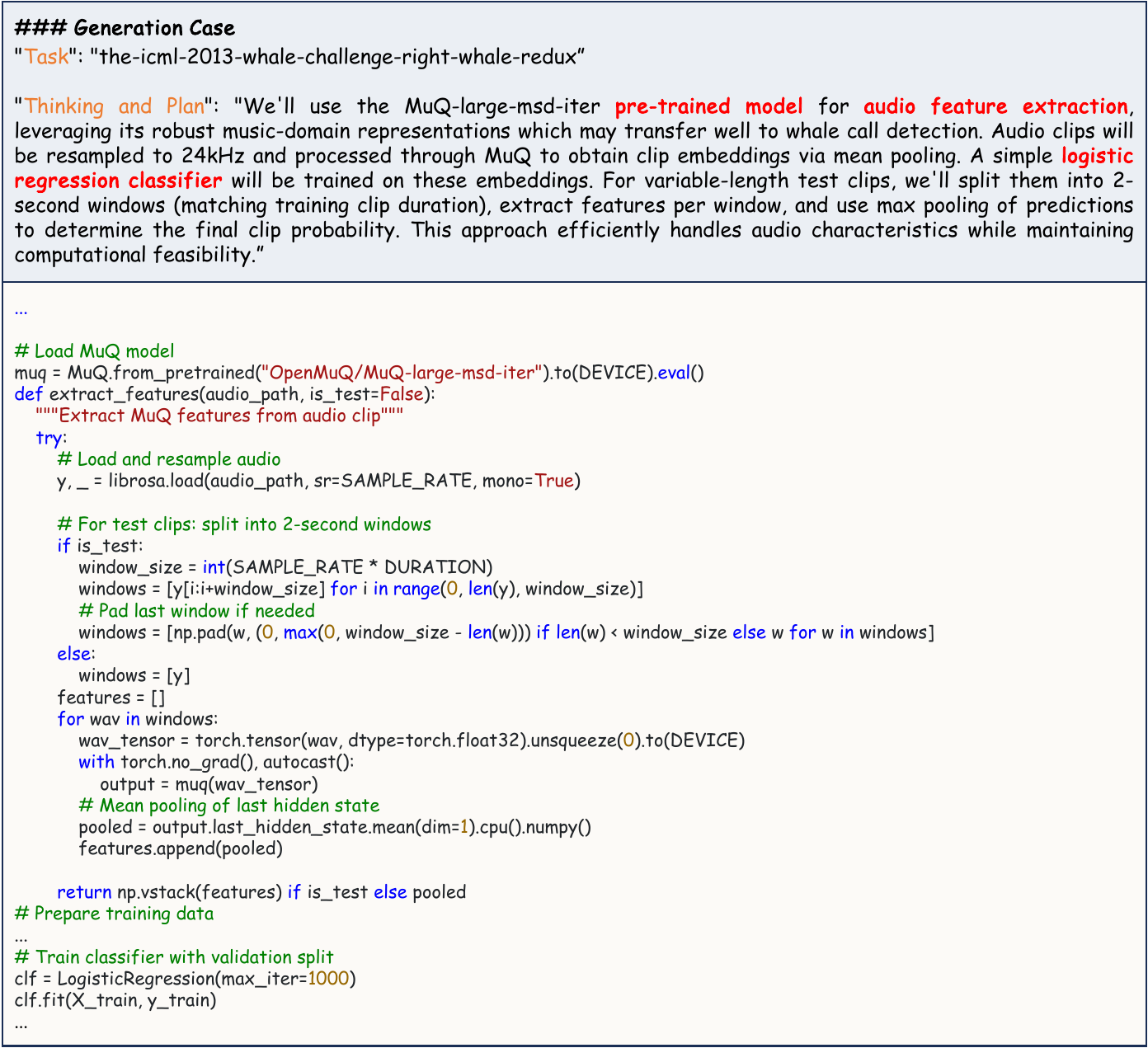}
    \vspace{-0.5em}
    \caption{\textbf{Case of code generation.}}
    \label{fig:case1}
\end{center}
\end{figure}

\begin{figure}[thbp]  
\begin{center}
    \vspace{-0.5em}
    \includegraphics[width=1.0\linewidth]{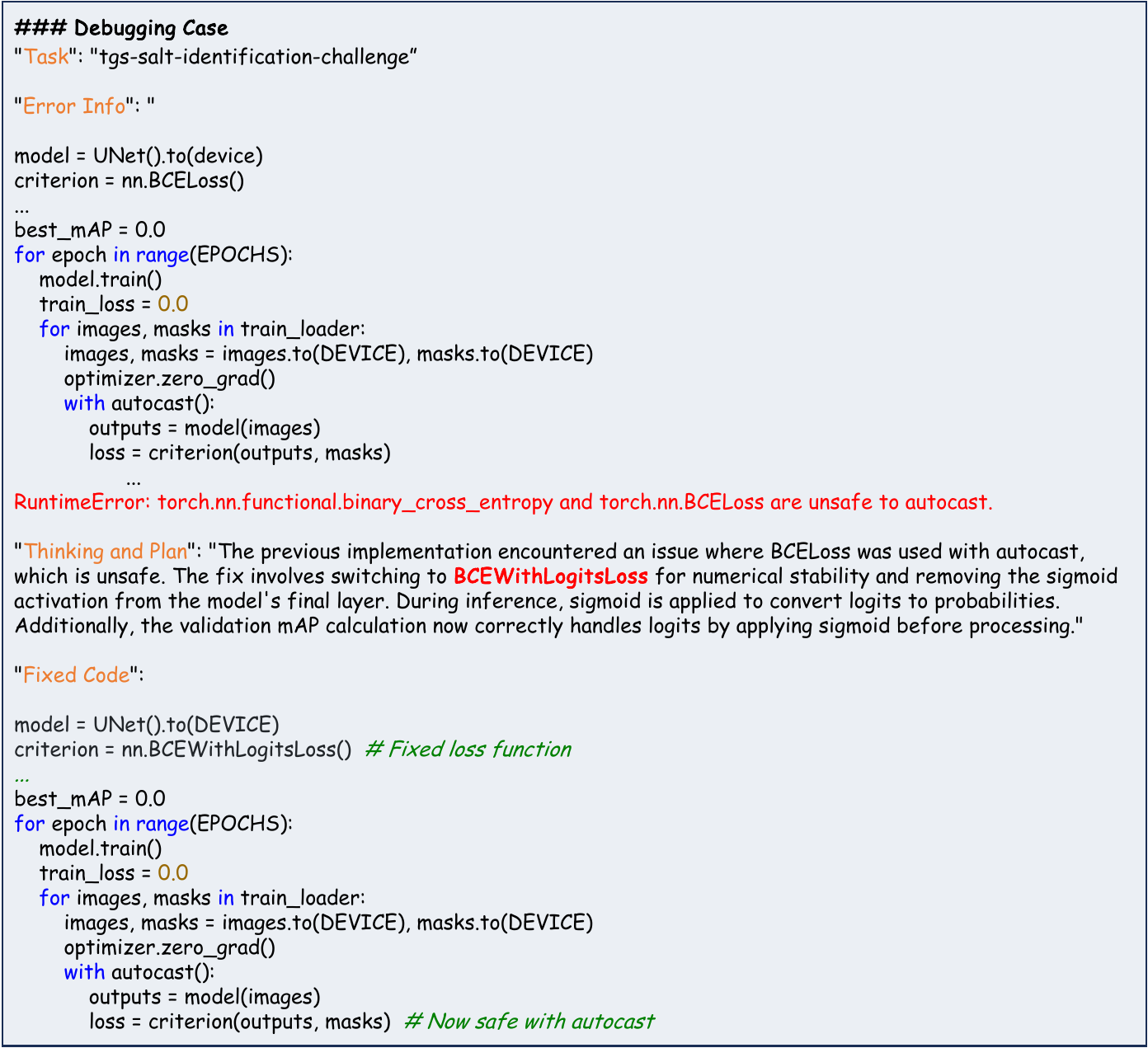}
    \vspace{-0.5em}
    \caption{\textbf{Case of debugging process.}}
    \label{fig:case2}
\end{center}
\end{figure}

\begin{figure}[thbp]  
\begin{center}
    \vspace{-0.5em}
    \includegraphics[width=1.0\linewidth]{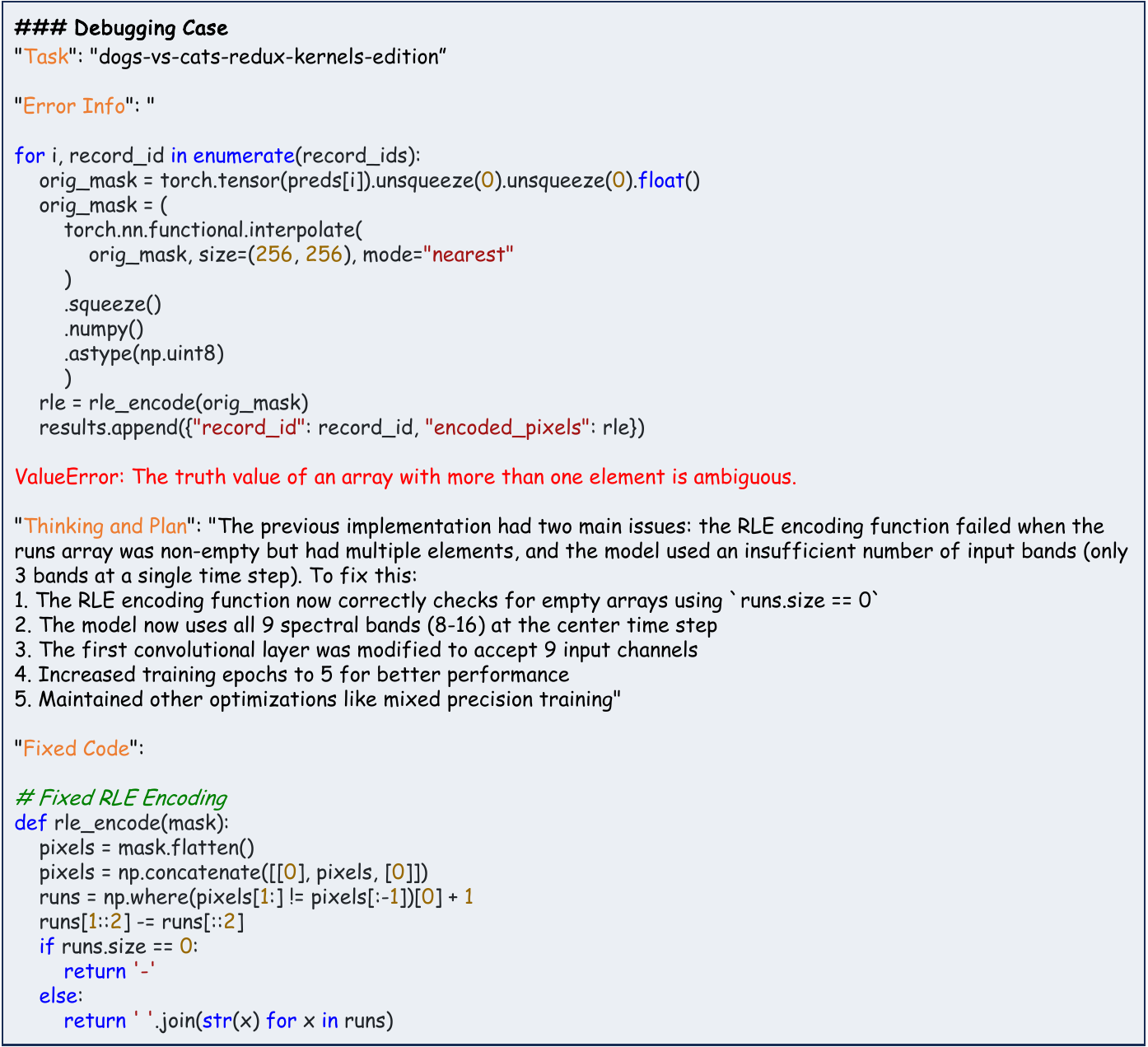}
    \vspace{-0.5em}
    \caption{\textbf{Case of debugging process.}}
    \label{fig:case3}
\end{center}
\end{figure}

\begin{figure}[thbp]  
\begin{center}
    \vspace{-0.5em}
    \includegraphics[width=1.0\linewidth]{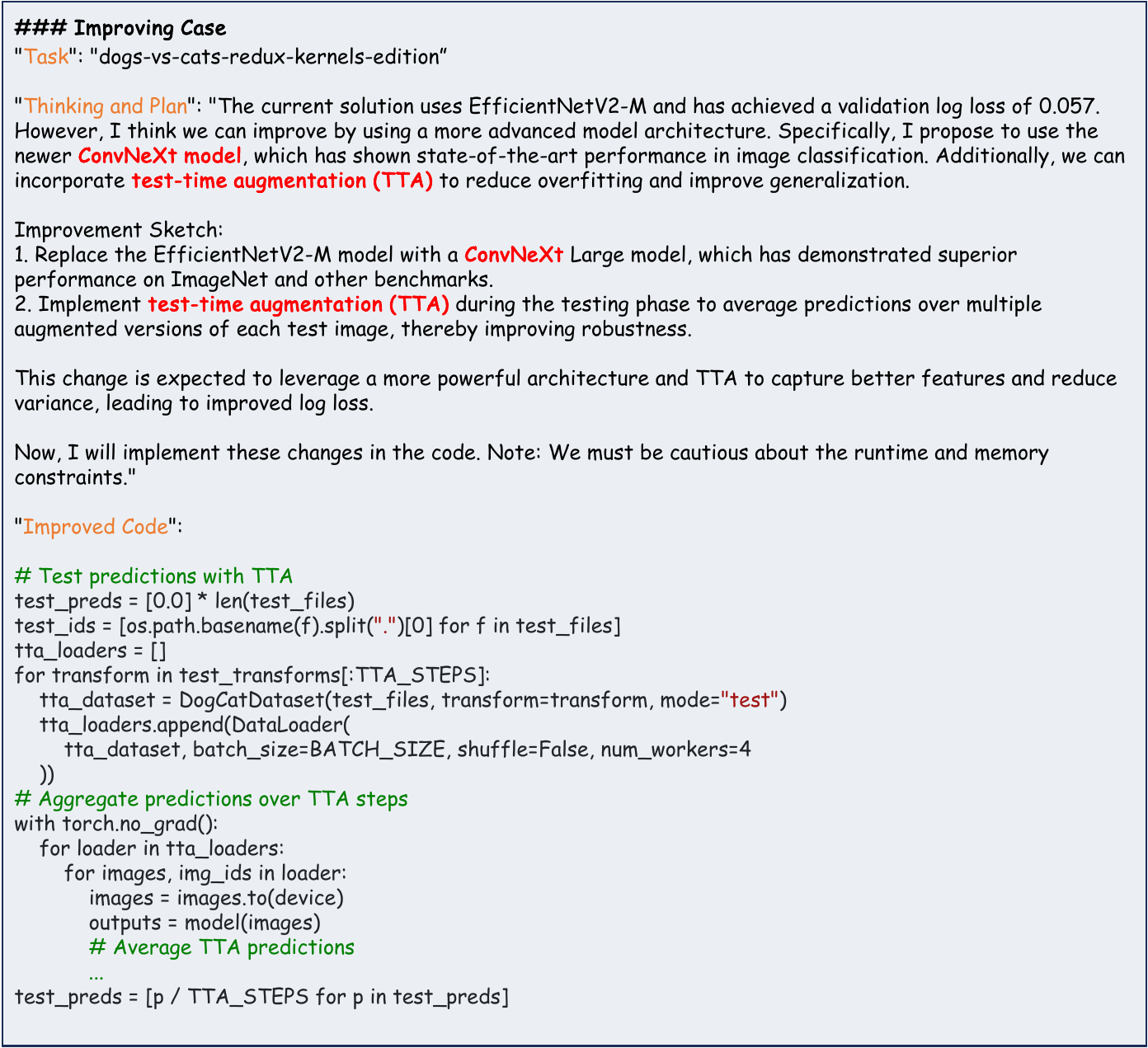}
    \vspace{-0.5em}
    \caption{\textbf{Case of improving process.}}
    \label{fig:case4}
\end{center}
\end{figure}

As shown in Fig~\ref{fig:case1}, \ref{fig:case2}, \ref{fig:case3}, and \ref{fig:case4}, we provide several representative cases that demonstrate \sname's autonomous ML engineering capabilities across different domains. These cases showcase the complete workflow from problem understanding to solution implementation, highlighting the agent's reasoning process, code generation, and performance optimization strategies.

\end{document}